\documentclass{bmvc2k}

\def\methodName{DisPositioNet}

\usepackage{amsmath,amssymb} % define this before the line numbering.
\usepackage{booktabs}
\usepackage{capt-of}
\usepackage{float}
\usepackage{tablefootnote}

\usepackage[capitalize]{cleveref}
\usepackage{multirow}
\usepackage{multicol}
\usepackage{bm}
\usepackage{bbding}
\usepackage{xspace}
\newcommand{\xpm}[1]{{\tiny$\pm#1$}}

\crefname{section}{Sec.}{Secs.}
\Crefname{section}{Section}{Sections}
\Crefname{table}{Table}{Tables}
\crefname{table}{Tab.}{Tabs.}
\newcommand\blfootnote[1]{%
  \begingroup
  \renewcommand\thefootnote{}\footnote{#1}%
  \addtocounter{footnote}{-1}%
  \endgroup
}

\title{\hspace{20pt}DisPositioNet: Disentangled Pose and \\\hspace{20pt} Identity in Semantic Image Manipulation}

\addauthor{Azade Farshad$^\ast$}{azade.farshad@tum.de}{1,2} % \thefootnote{}
\addauthor{Yousef Yeganeh$^\ast$}{y.yeganeh@tum.de}{1} % \thefootnote{}
\addauthor{Helisa Dhamo$^\ddagger$}{helisa.dhamo@huawei.com}{3} %helisa.dhamo@huawei.com
\addauthor{Federico Tombari}{tombari@in.tum.de}{1,4} %tombari@in.tum.de
\addauthor{Nassir Navab}{nassir.navab@tum.de}{1,5} %nassir.navab@tum.de

\addinstitution{
 Technical University of Munich \\
 Germany
}

\addinstitution{
 Munich Center for Machine Learning\\
 Germany
}

\addinstitution{Huawei Noah's Ark Lab \\
United Kingdom
}

\addinstitution{
 Google \\
 Switzerland
}

\addinstitution{Johns Hopkins University\\
USA
}

\runninghead{Farshad, Yeganeh, Dhamo, Tombari, Navab}{\methodName{}}

\begin{document}

\maketitle
\blfootnote{$^\ast$ Equal Contribution.}
\blfootnote{$^\ddagger$ The work was mostly done while Helisa was at TU Munich.}
%%%%%%%%% ABSTRACT
\begin{center}
\centering
\vspace{-0.8cm}
    \includegraphics[width=\textwidth]{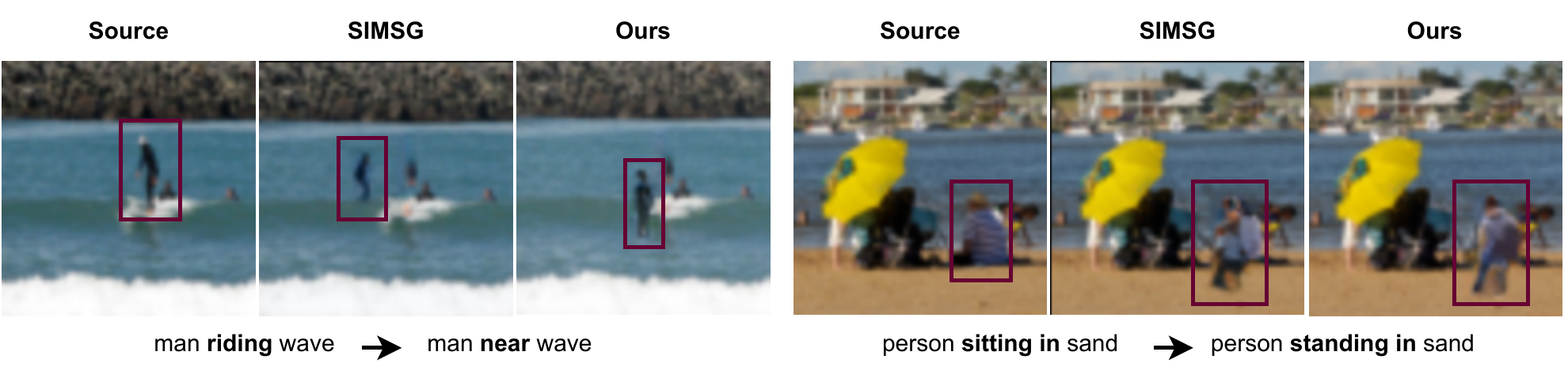} \vspace{-0.6cm}
    \captionof{figure}{Our method is able to preserve the object features required for image manipulation and generate more realistic objects compared to SIMSG \cite{dhamo2020semantic}.}
    \label{fig:compare}
\end{center}
\begin{abstract}
 Graph representation of objects and their relations in a scene, known as a scene graph, provides a precise and discernible interface to manipulate a scene by modifying the nodes or the edges in the graph. Although existing works have shown promising results in modifying the placement and pose of objects, scene manipulation often leads to losing some visual characteristics like the appearance or identity of objects. In this work, we propose \methodName{}, a model that learns a disentangled representation for each object for the task of image manipulation using scene graphs in a self-supervised manner. Our framework enables the disentanglement of the variational latent embeddings as well as the feature representation in the graph. In addition to producing more realistic images due to the decomposition of features like pose and identity, our method takes advantage of the probabilistic sampling in the intermediate features to generate more diverse images in object replacement or addition tasks. The results of our experiments show that disentangling the feature representations in the latent manifold of the model outperforms the previous works qualitatively and quantitatively on two public benchmarks. \textbf{Project Page}: \url{https://scenegenie.github.io/DispositioNet/}
\end{abstract}

%%%%%%%%% BODY TEXT
\section{Introduction}
\label{sec:intro}
Image manipulation is a task of interest in computer vision, which consists of the partial synthesis, change, or removal of the content in a given image. In recent years, this task has been explored using deep generative models, in particular utilizing Generative Adversarial Networks (GANs) \cite{goodfellow2014generative}. There have been different approaches towards image manipulation, in which the interface used to induce these changes by a user is an important choice. The utilization of segmentation maps for image modification \cite{hong2018learning,ntavelis2020sesame} requires direct manipulation of a semantic segmentation map at the pixel level. Recently, motivated by a more user-friendly interface, SIMSG \cite{dhamo2020semantic} proposed a semantic manipulation framework using scene graphs. Scene graphs define a scene by considering the objects in the scene as the nodes in the graph, and the edges as the relationships between the objects. In semantic image manipulation using scene graphs, the user simply needs to change the nodes or edges in a graph that represents the scene. The manipulation of scenes in SIMSG \cite{dhamo2020semantic} is performed by masking specific parts of data based on the manipulation mode, e.g., the object features or the bounding box information.

Despite the encouraging results, this model comes with a pitfall that the learned object features used for manipulation are intertwined, i.e., they encode both the pose and appearance features simultaneously. This becomes particularly evident when we want to preserve one of the aspects of an object while changing the other. For instance, in \cref{fig:compare} we observe a relationship change setup. When the \texttt{man} changes from \texttt{riding} to \texttt{near} the \texttt{wave} (left), or from \texttt{sitting} to \texttt{standing} in the \texttt{sand} (right), SIMSG \cite{dhamo2020semantic} (middle column) will lose some visual features of the man, in the process of adapting to the new pose, i.e. change of body shape or outfit color. 

In this work, we propose \textbf{DisPositioNet}, a network for \textbf{Dis}entangled \textbf{Pos}e and \textbf{I}dentity in Semantic Image Manipula\textbf{tion}, which disentangles the object features using a self-supervised variational approach by employing two branches for encoding the pose and appearance features in the latent space. We hypothesize that, by disentangling the features in the image manipulation framework, the model would preserve features more reliably, and therefore generate more meaningful results. To disentangle the features further and make the extracted features from the scene graph more compatible with the variational embedding, we propose DSGN, a disentangled scene graph neural network for disentangled feature extraction from the scene graphs. We evaluate our model on standard benchmarks for image manipulation (Visual Genome \cite{krishna2017visual}, and Microsoft COCO \cite{lin2014microsoft}), showing superior performance compared to the baseline \cite{dhamo2020semantic} both quantitatively and qualitatively. The qualitative results show that our proposed method specifically outperforms SIMSG \cite{dhamo2020semantic} in cases where the appearance of the object should be preserved while changing its pose. 

To summarize our contributions, we propose: 1) a self-supervised approach for disentanglement of pose and appearance for semantic image manipulation, that does not require label information for the disentanglement task, 2) a disentangled scene graph neural network, 3) a variational latent representation that provides higher diversity in image manipulation, 4) superior quantitative and qualitative performance compared to the state of the art on two public benchmarks. The source code of this work is provided in the supplementary material, and it will be publicly released upon its acceptance.

\section{Related Work}
\label{sec:rel}

\paragraph{Scene Graphs}
Scene graphs define a directed graph representation that describes an image \cite{johnson15}, where objects are the nodes and their relationships are the edges. 
A broad line of works explores the generation of scene graphs from an image \cite{herzig2018mapping,qi2018attentive,zellers2018neural,li2017scene,newell2017pixels,tang2020unbiased,suhail2021energy} and recently also point clouds~\cite{3DSSG2020,wu2021scenegraphfusion}. The task boils down to identifying the underlying objects in a scene and their visual relationships. A diverse set of approaches has been explored for this purpose, such as iterative message-passing \cite{xu2017scenegraph}, decomposition of the graph into sub-graphs \cite{li2018factorizable} and attention mechanisms \cite{yang2018graph}. Recently, SceneGraphGen \cite{garg2021unconditional} explored this task unconditionally by learning an auto-regressive model on scene graphs. Scene graphs have shown to be a powerful alternative in conditional scene generation \cite{johnson2018image,luo2020end,graph2scene2021}, and manipulation \cite{dhamo2020semantic}, which we will review as it follows.

\paragraph{Image Generation}
The recent advances in image generation, for the most part, emerged from Generative Adversarial Networks \cite{goodfellow2014generative} and diffusion models \cite{nichol2021improved}. In particular, the community has explored conditional variants \cite{mirza2014conditional} which enable image generation conditioned on various modalities.
Pix2Pix \cite{isola2017image} represents a model for general translation between different image domains. Further, CycleGAN \cite{zhu2017unpaired} attempts this task by relaxing the need for image pairs for training. Other works \cite{karras2020analyzing,karras2021alias} explore unconditional generation, typically focused on a specific domain, such as faces. A line of methods \cite{chen2017photographic,wang2018high,park2019SPADE} propose semantic image generation, where an image results from an input semantic map.
Other works propose image generation from layout \cite{zhao2018image,Sun_2019_ICCV}, as a set of bounding boxes and class labels for each scene instance. More related to ours are methods that generate an image conditioned on a scene graph \cite{johnson2018image,ashual2019specifying,farshad2021_MIGS,ivgi2021scene}, where the layout arises as an intermediate step to translate the graph structure into image space. Johnson et al. \cite{johnson2018image} introduced Sg2im, the first method that tackles this task supervised via a combined object-level and image-level GAN loss. Following work further improve the performance in this challenging task by utilizing per-object neural image features to increase diversity \cite{ashual2019specifying}, exploiting meta-learning to better learn the highly diverse datasets (MIGS) \cite{farshad2021_MIGS}, and employing contextual information to refine the layout (CoLoR) \cite{ivgi2021scene}.

\paragraph{Interactive Image Manipulation} This task represents a form of partial image generation, which usually comes with a user interface to indicate the subject of change \cite{li2022cross}.
Early works perform scene-level image editing in a hand-crafted manner, which replaces some image parts with sample patches from a database \cite{hu2013patchnet}. One form of manipulation is image inpainting, where a user can indicate a mask for removing and automatically filling an image area \cite{pathak2016context}, that can be further extended with semantics \cite{yeh2017semantic} or edges \cite{Yu_2019_ICCV,nazeri2019edgeconnect} to guide the missing area.
%Hierarchical Manipulation 
Hong et al. \cite{hong2018learning} employ a learned model on a semantic layout representation, in which the user can make changes in the image by adding, moving, or removing bounding boxes. 
SESAME \cite{ntavelis2020sesame} allows the user to draw a mask with semantic labels on an image to indicate the category of the changed pixels. Similarly, in EditGAN \cite{ling2021editgan} the user can modify a detailed object part segmentation map to alter object appearance. 
SIMSG \cite{dhamo2020semantic} explores scene graphs as the interface, where the user can make changes in the nodes or edges of a graph to manipulate the image. 
Recently, Su et al.  \cite{su2021fully} proposed an improvement to this model by relying on masks instead of bounding boxes for the object placement. 
Different from these models, we want to model an object representation with disentangled appearance and pose.
\begin{figure}[tb]
\centering
    \includegraphics[width=\textwidth]{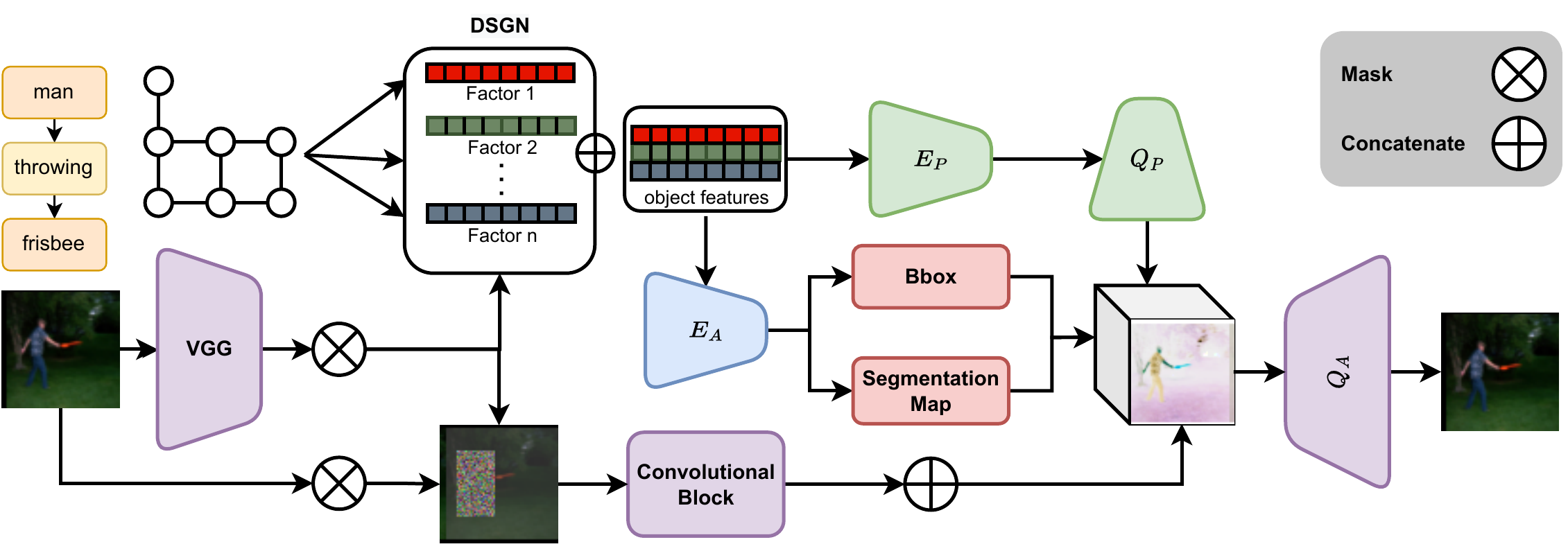}
    \caption{\textbf{\methodName{} Overview.} Our disentangled image manipulation framework performs by disentangling the graph representations, as well as the variational embeddings through learning the feature transformations.}
    \label{fig:method}
    %\par\bigskip}]
\end{figure}
\paragraph{Disentangled Representation Learning} Learning disentangled representation has been explored in many works using variational autoencoders \cite{esser2018variational,tatro2020unsupervised}, e.g., for changing the digit and handwriting in the MNIST dataset. Initial works on disentangled representation learning \cite{chen2016infogan} focused on variational mutual information maximization or decreasing the channel capacity of the variational autoencoder \cite{higgins2016beta}. Some recent works focus on disentanglement using deformable networks \cite{shu2018deforming,xing2019unsupervised}, contrastive learning \cite{bai2021contrastively,parmar2021dual} or disentangling identity and pose for face manipulation \cite{zeno2019ip}. Many of these works require labeled information for conditioning the model on the specific attributes for the disentangling. They also disentangle the data without knowing which feature belongs to which factor. Recently, \cite{skafte2019explicit} was proposed as an unsupervised way to disentangle pose and appearance. They used two branches to predict the image's transformation parameters and apply the learned transformation to the appearance features.
%Elastic InfoGAN \cite{ojha2019elastic},
Disentanglement in graph neural networks (GNN) has been previously explored in \cite{ma2019disentangled,yang2020factorizable} where the graph features are divided into different factors that help disentangle the latent representation. These approaches, however, operate on regular graphs, which only consider neighboring nodes when computing the features. In contrast, GNNs designed for scene graphs also modulate edge features into the GNN network.

\section{Method}
\label{sec:method}
Our goal is to learn a disentangled representation for the appearance and pose of the objects in the latent space for the semantic image manipulation task to preserve the features of specific attributes. As it follows, we first discuss the semantic image manipulation framework. Then, we describe our proposed disentangled graph model and our variational disentanglement approach in detail. \cref{fig:method} shows an overview of our method.

\paragraph{Semantic Image Manipulation}\label{sec:method_simsg}
 Given an image $I$ and its corresponding scene graph $\mathcal{G} = \{\mathcal{O}, \mathcal{R}\}$, where $\mathcal{O}$ is the set of objects (nodes) in the scene and $\mathcal{R}$ represents the set of relationships (edges) between the objects, the goal is to obtain a modified image $I^{*}$ based on an altered version of the scene graph $\mathcal{G}^{*}$. Inspired by SIMSG~\cite{dhamo2020semantic} we formulate this image manipulation task via a reconstruction proxy objective, such that we do not need to rely on image pairs with changes for the training. To enable control on specific object attributes, the semantic graph representation is extended to obtain an augmented graph, where each node contains a semantic class embedding, a bounding box $x$, and a neural visual feature $z_I$. During training, object regions in the image, visual features, or bounding boxes are randomly masked using a noise vector, and the model's objective is to reconstruct the masked parts using the information from the scene graph $\mathcal{G}$ and the remaining regions in the image. $\mathcal{G}$ is defined as a set of triplets $\mathcal{G}_i = (s_i, r_i, o_i)$, where $s_i, r_i, o_i$ are the subject, predicate and object respectively. Each object in the graph belongs to a class of object categories $\mathcal{C} = \{c_1,c_2, ..., c_n\}$. Image features $z_I$ are extracted from the input image using a pre-trained classifier network such as VGG16 \cite{simonyan2014very}. The graph triplets are fed to a scene graph neural network (SGN) for message passing between the nodes. $z_\mathcal{G}$ is obtained from SGN with parameters $\Phi$, which processes the scene graph $\mathcal{G}$, the input bounding boxes $x$, and visual features $z_I$.

To disentangle the features based on the pose and appearance, we harness two encoder networks, namely $E_A$ and $E_P$, that receive the per-object features $z_\mathcal{G}$ as input and produce appearance features $z_{\mathcal{G}A}$ and pose features $z_{\mathcal{G}P}$ respectively. The object bounding boxes and pseudo-segmentation maps are predicted utilizing two networks that receive $z_{\mathcal{G}A}$ as input. Further, the scene layout $z_l$ is constructed by projecting the appearance features $z_{\mathcal{G}A}$ of each object in the image space, in the regions indicated by the respective predicted bounding boxes and segmentation maps. We further employ a pose decoder network $Q_P$ to predict a set of transformation parameters $\gamma$. These parameters are used to construct a transformation function $\tau$, which is applied to the pooled object features from the scene layout $z_l$. The object feature pooling is performed by cropping $z_l$ using the bounding boxes $x$ and applying the per-object transformations on the cropped feature vectors. Finally, the reconstructed image $\tilde I$ is generated by passing the transformed layout $\tau(z_l)$ to the image decoder network $Q_A$.

\paragraph{Disentangled Graph Neural Network}\label{sec:method_disengcn}
One main limitation in the SIMSG formulation is that the object features extracted by the SGN are entangled. Therefore, to increase the disentanglement between pose and appearance even further, we propose using a disentangled graph neural network. Inspired by \cite{ma2019disentangled}, we propose \textit{DSGN}, a disentangled scene graph network. DSGN not only considers the nodes in the graph as in \cite{ma2019disentangled}, but it also combines the edge features (here predicates) in the disentangled feature extraction, as this provides crucial information for the task at hand. Our network is thus adapted for triplets of the form $\mathcal{G}_i = (s_i, r_i, o_i)$. The DSGN utilizes disentangled convolutional layers combined with neighborhood routing mechanism \cite{sabour2017dynamic} for projection of the features into different subspaces. The neighborhood routing mechanism actively distinguishes the latent factor that could have caused the edge between a node and its neighbors. This would assign the neighbor to another channel to extract the features for that specific factor. The DSGN receives the triplets $\mathcal{G}_i = (s_i, r_i, o_i)$ as input, where $o_i, s_i \in \mathcal{O}$ and $r_i \in \mathcal{R}$. Each layer $e$ in the DSGN represents a function $f_e(.)$ which applies the edge features to the nodes in the graph and their neighbours:
\begin{equation}
    (\alpha_{ij}^{(t+1)}, \rho_{ij}^{(t+1)}, \beta_{ij}^{(t+1)}) = f_e\left(
    \nu_i^{(t)}, \,\rho_{ij}^{(t)}, \,\nu_j^{(t)}
    \right) ,
\end{equation}
with $\nu_i^{(0)} = o_i$, where $t$ represents the layers of the DSGN, which has a total of $T$ layers. The input is first processed by a Sparse Input Layer \cite{feng2017sparse} to decompose the node features $\nu_i^{(t)}$ into $k$ factors. The $k$ node features are then passed through $k$ separate neighbourhood routing layers. Finally, the object features $\nu_i^{(T)}$ are computed by concatenating the object features from all factors $k$.

\paragraph{Disentangled Variational Embedding}\label{sec:method_disenVAE}
 Our model is enforced to disentangle the perspective and appearance features in the latent embedding, through modelling and predicting the transformation in the features, based on \cite{skafte2019explicit}. The variational embedding disentanglement happens by employing two encoder and two decoder networks. The object features $z_\mathcal{G}$ are passed to the variational encoders $E_A, E_P$, that are composed of two subnetworks that model the mean $\mu$ and variance $\sigma$ of the data. They both output the latent representation $z$, obtained by applying the reparameterization trick $z = \mu + \sigma \epsilon$, where $\epsilon$ is a random noise vector.

$E_A$ encodes the appearance features, while $E_P$ encodes the perspective information. The transformation $\gamma$ for each object is predicted by a simple MLP network $Q_P$, which is then applied to the pooled object patches from the scene layout $z_l$. The intention behind predicting and applying the transformation $\gamma$ by $Q_P$ is to separate the pose information in the pose branch and enforce the model to only learn the appearance features in $E_A$. Finally, the image is reconstructed from the scene layout $z_l$ by $Q_A$, which is the SPADE \cite{park2019SPADE} generator here.

We define the affine transformation function $\tau$ given the input $z$ as follows:
\iffalse
\begin{equation}
\label{eq:transform_gen}
    \tau_\gamma (z) = \begin{bmatrix}
\gamma_{11} & \gamma_{12} & \gamma_{13}\\
\gamma_{21} & \gamma_{22} & \gamma_{23}\\
\gamma_{31} & \gamma_{32} & \gamma_{33}
\end{bmatrix}
\begin{bmatrix}
z_a \\
z_b \\
1
\end{bmatrix}
\end{equation}
which can be then defined as:
\fi
\begin{equation}
\begin{split}
  \tau_{\gamma,\text{affine}} (z) &= \begin{bmatrix}
\cos(\alpha) & -\sin(\alpha)\\
\sin(\alpha) & \cos(\alpha)
\end{bmatrix} \begin{bmatrix}
1 & m\\
0 & 1
\end{bmatrix} \begin{bmatrix}
\delta_{z_a} & 0\\
0 & \delta_{z_b}
\end{bmatrix} + %&\quad + 
\begin{bmatrix}
t_{z_a} \\
t_{z_b}
\end{bmatrix}
\end{split}
\end{equation}
where $\alpha$ is the rotation angle, $m$ is the shear value, $\delta_{z_a}$ and $\delta_{z_b}$ are the scaling factors and $t_{z_a}$, $t_{z_a}$ are the translation parameters. These parameters, defined by $\gamma$ are modelled by an MLP represented by $Q_P$ that outputs these values per object.

\paragraph{Objective Functions}
The loss terms used for training our model are a combination of original losses used in \cite{dhamo2020semantic} and variational terms. The generative adversarial objective is:
\begin{equation}
  \mathcal{L}_\text{GAN} = \mathop{\mathbb{E}}_{q\sim p_{\textrm{data}}} \log D(q) + \mathop{\mathbb{E}}_{q\sim p_{\textrm{g}}} \log(1 - D(q)),
\end{equation}
where $p_g$ denotes the distribution of fake / generated images or object patches, $p_{data}$ is the distribution of the ground truth images or objects, and $q$ defines the input to the discriminator network $D$ sampled from the ground truth or generated data distributions. In addition to the global image discriminator $D_{image}$, an object discriminator $D_{obj}$ is used for cropped patches of objects in the image to improve the appearance and realism of the objects. The bounding box prediction loss is defined as $\mathcal{L}_{bbox} = \lambda_b \left\lVert x_i - \hat{x}_i \right\rVert_1^1$, while the generative objective is:

\begin{equation}
  \begin{split}
      \mathcal{L}_\text{generative} & =  \lambda_g \min_G \max_D \mathcal{L}_\text{GAN,image} + %\\  &\quad + 
      \lambda_o \min_G \max_D \mathcal{L}_\text{GAN,obj} \\  &\quad +
      \lambda_a \mathcal{L}_\text{aux,obj} + \mathcal{L}_{rec} + %\\ &\quad +
      \lambda_p \mathcal{L}_p + \lambda_f \mathcal{L}_f,
  \end{split}
\end{equation}
where $\lambda_g$, $\lambda_o$, $\lambda_a$ are the constant weight multipliers. $\mathcal{L}_\text{aux,obj} $ is an auxiliary object classifier loss~\cite{Odena2017ConditionalIS}, and $\mathcal{L}_p,\mathcal{L}_f$ are respectively the perceptual and GAN feature loss terms borrowed from the SPADE generator \cite{park2019SPADE}, and $L_{rec} = \| I - \tilde{I} \|_1$ is the image reconstruction loss.

The variational objective for feature disentanglement tries to minimize the evidence lower bound (ELBO):

\begin{equation}
\begin{split}
    \mathcal{L}_{var} &= \mathbb{E}_{q_A,q_P} [log(p(I|z_{\mathcal{G}A}, z_{\mathcal{G}A}))] %\\  &\quad
    - D_{\mathbb{K}\mathbb{L}} (q_P(z_{\mathcal{G}P}|I)||p(z_{\mathcal{G}P})) \\  &\quad
    - \mathbb{E}_{q_P}[D_{\mathbb{K}\mathbb{L}} (q_A(z_{\mathcal{G}A}|I)||p(z_{\mathcal{G}A}))].
      \end{split}
\end{equation}

Then, the final objective becomes:
\begin{equation}
    \mathcal{L}_{total} = \mathcal{L}_{var} + \mathcal{L}_\text{generative} + \mathcal{L}_{bbox}
\end{equation}

\begin{table}[htb]
    \centering
    \footnotesize %
        \caption{\textbf{Image reconstruction on Visual Genome.} We compare the results of our method to previous works using ground truth (GT) and predicted scene graphs. In the experiments denoted by (Generative), the whole input image is masked. N/A: Not Applicable.}
        \resizebox{\columnwidth}{!}{
    \begin{tabular}{l@{\hskip 16pt} c@{\hskip 26pt} c@{\hskip 12pt} c@{\hskip 12pt} c@{\hskip 12pt} c@{\hskip 12pt} c@{\hskip 12pt} c@{\hskip 12pt} c@{\hskip 10pt}}
    \toprule
    \multirow{2}{*}{Method} & \multirow{2}{*}{Decoder} & \multicolumn{5}{c}{All pixels} & \multicolumn{2}{c}{RoI only} \\
    \cmidrule(l{1pt}r{9pt}){3-7} \cmidrule(l{1pt}r{3pt}){8-9}
     & & MAE $\downarrow$ & SSIM $\uparrow$ & LPIPS $\downarrow$ & FID $\downarrow$ & IS $\uparrow$ & MAE $\downarrow$ & SSIM $\uparrow$ \\
    \midrule
    \multicolumn{9}{c}{Generative, GT Graphs} \\
    \midrule
    ISG~\cite{ashual2019specifying}  & Pix2pixHD & $46.44$ & $28.10$ & $0.32$ & 58.73 & $6.64$\xpm{0.07} & N/A & N/A \\
    SIMSG~\cite{dhamo2020semantic} & SPADE & $41.88$ & $34.89$ & $0.27$ & $44.27$ & $7.86$\xpm{0.49} & N/A & N/A \\
    \methodName{} (Ours) & SPADE & $\bm{41.62}$ & $\bm{35.30}$ & $\bm{0.26}$ & $\bm{40.75}$ & $\bm{7.93}$\xpm{0.36} & N/A & N/A \\
    \midrule \midrule
    \multicolumn{9}{c}{GT Graphs} \\
    \midrule
    Cond-sg2im \cite{johnson2018image} & CRN & $14.25$ & $84.42$ & $0.081$ & $13.40$ & $11.14$\xpm{0.80} & $29.05$ & $52.51$\\
    SIMSG \cite{dhamo2020semantic} & SPADE &  $8.61$ & $87.55$ & $0.050$ & $\bm{7.54}$ & $\bm{12.07}$\xpm{0.97} & $\bm{21.62}$ & $\bm{58.51}$\\
    \methodName{} (Ours) & SPADE &  $\bm{8.41}$ & $\bm{87.56}$ & $\bm{0.048}$ & $7.66$ & $11.65$\xpm{0.58} & $21.76$ & $58.18$\\
    \midrule
    \multicolumn{9}{c}{Predicted Graphs} \\
    \midrule
    SIMSG \cite{dhamo2020semantic} & SPADE & $13.82$ & $83.98$ & $0.077$ & $16.69$ & $10.61$\xpm{0.37} & $28.82$ & $49.34$ \\
    \methodName{} (Ours) & SPADE & $\bm{9.39}$ & $\bm{86.91}$ & $\bm{0.052}$ & $\bm{14.42}$ & $\bm{10.69}$\xpm{0.33} & $\bm{25.40}$ & $\bm{51.85}$ \\
    \bottomrule
\end{tabular} }%
\label{tab:vg}
\end{table}

\begin{figure}[htb]
    \centering
    \includegraphics[width=\textwidth]{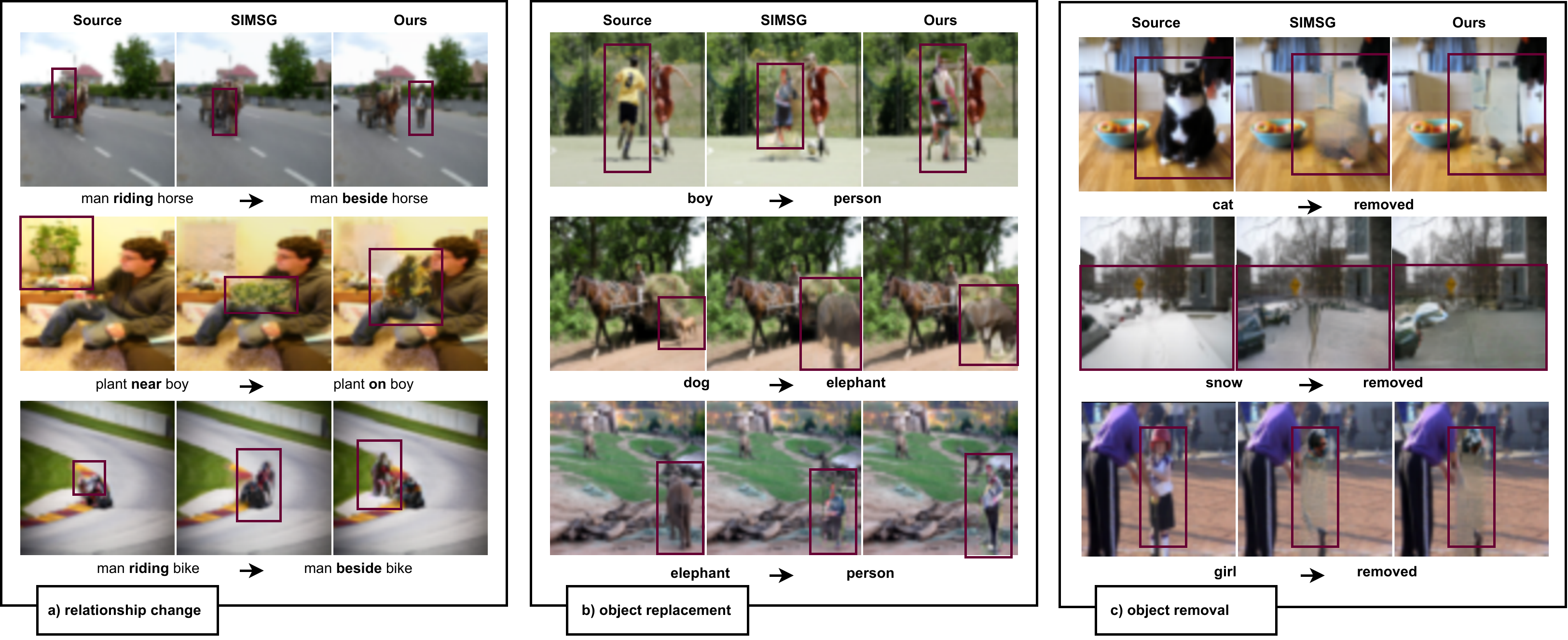}
    \caption{\textbf{Qualitative comparison to SIMSG \cite{dhamo2020semantic} on VG.} It can be seen that in a) the plant has a more realistic appearance and a more similar shape to the original object, the same applies to b) where the boy and the elephant are changed to person. For the object removal in c) there are some artifacts visible after the removal of the cat and snow, however the images generated by our method do not have these artifacts and look more realistic.}
    \label{fig:vg_qualitative}
\end{figure}

\section{Experiments}
\label{sec:results}
In this section, we first discuss our framework's setup, including the hyperparameters and the metrics. Then we ablate the different components of our model, and report the quantitative and qualitative results of our experiments compared to previous work. We also provide the results of our performed user study. Finally, we discuss the results and the limitations of our work.
\subsection{Experimental Setup}
%\paragraph{Datasets}
We evaluate our method on Visual Genome (VG) \cite{krishna2017visual} and Microsoft COCO \cite{lin2014microsoft} datasets which are commonly used in the image generation using scene graphs literature. The qualitative results on COCO are included in the supplementary material due to the space limitations.
\paragraph{Evaluation Metrics}
To evaluate the quality of generated images by our model, we use common similarity metrics used for GANs such as Inception Score (IS) \cite{salimans2016improved}, Frechet Inception Distance (FID) \cite{heusel2017gans}, structural similarity metric (SSIM) \cite{wang2004image}, Perceptual Similarity (LPIPS) \cite{zhang2018perceptual} and the Mean Absolute Error (MAE). We also measure the MAE and SSIM for the Region of Interest (RoI) where the change or reconstruction happens.

\paragraph{Implementation Details}
All models were trained on $64 \times 64$ images with batch size of $32$. The visual feature extraction model is a VGG-16 pretrained on ImageNet. The learning rate for all models is $2e-4$, and the disentangling factor $k$ in the Disentangled SGN is equal to $16$. All models were trained for $300k$ iterations on VG and COCO. The architectures of $E_P$, $E_A$, and $D_P$ are MLPs with two FC layers with 64 filters, 1 BN layer, and a LeakyReLU activation function. The decoder and discriminator architectures follow \cite{dhamo2020semantic}. The values of the hyperparameters were obtained empirically or based on previous works. The slight difference between the reported values and the ones in \cite{dhamo2020semantic} could be due to library version differences. We report the details of network architectures in the supplementary material.

\paragraph{Modification Modes}
During the testing phase, four modification modes are supported, i.e. relationship change, object replacement, object removal, and object addition. The model receives the source image, and the desired modification on the graph as input. Specific features are masked based on the modification mode and the target image is generated by the decoder. E.g., for relationship changes, the object features are retained while the bounding box features are masked. On the other hand, for object replacement, the object features are dropped while the bounding box features are preserved. 

\subsection{Results}
\paragraph{Quantitative Results} Quantitative evaluation of image manipulation methods on a real-world dataset is a difficult task due to the lack of paired source and target data. Therefore, following previous work \cite{dhamo2020semantic}, we evaluate our method based on the reconstruction quality. The input images are partially masked, and the goal of the model is to reconstruct the masked parts from the information in the scene graph. The results of our experiments are presented in \cref{tab:vg} and \cref{tab:coco}. In the generative mode, the whole image is masked, and the model is generated purely from the scene graph to evaluate its image generation performance. The models are given either ground truth graphs as input or predicted ones by a scene graph to image model \cite{li2018factorizable}. The results show that, the \methodName{} model outperforms the state-of-the-art in almost all metrics and scenarios. We also provide the results of our user study on the comparison between SIMSG and \methodName{} for different manipulation modes in \cref{tab:user_study}.The user study details are provided in the supplementary material.

\paragraph{Qualitative Results} Some qualitative results of our method on VG dataset are shown in \cref{fig:vg_qualitative}. As it can be seen, our proposed method is able to learn better feature representations and therefore generate more meaningful results. We also provide some qualitative examples on diversity in the supplementary material, and show that in contrast to SIMSG, our model is able to generate diverse images in terms of color and texture.

\begin{figure}[!htb]
    \centering
    \includegraphics[width=\textwidth]{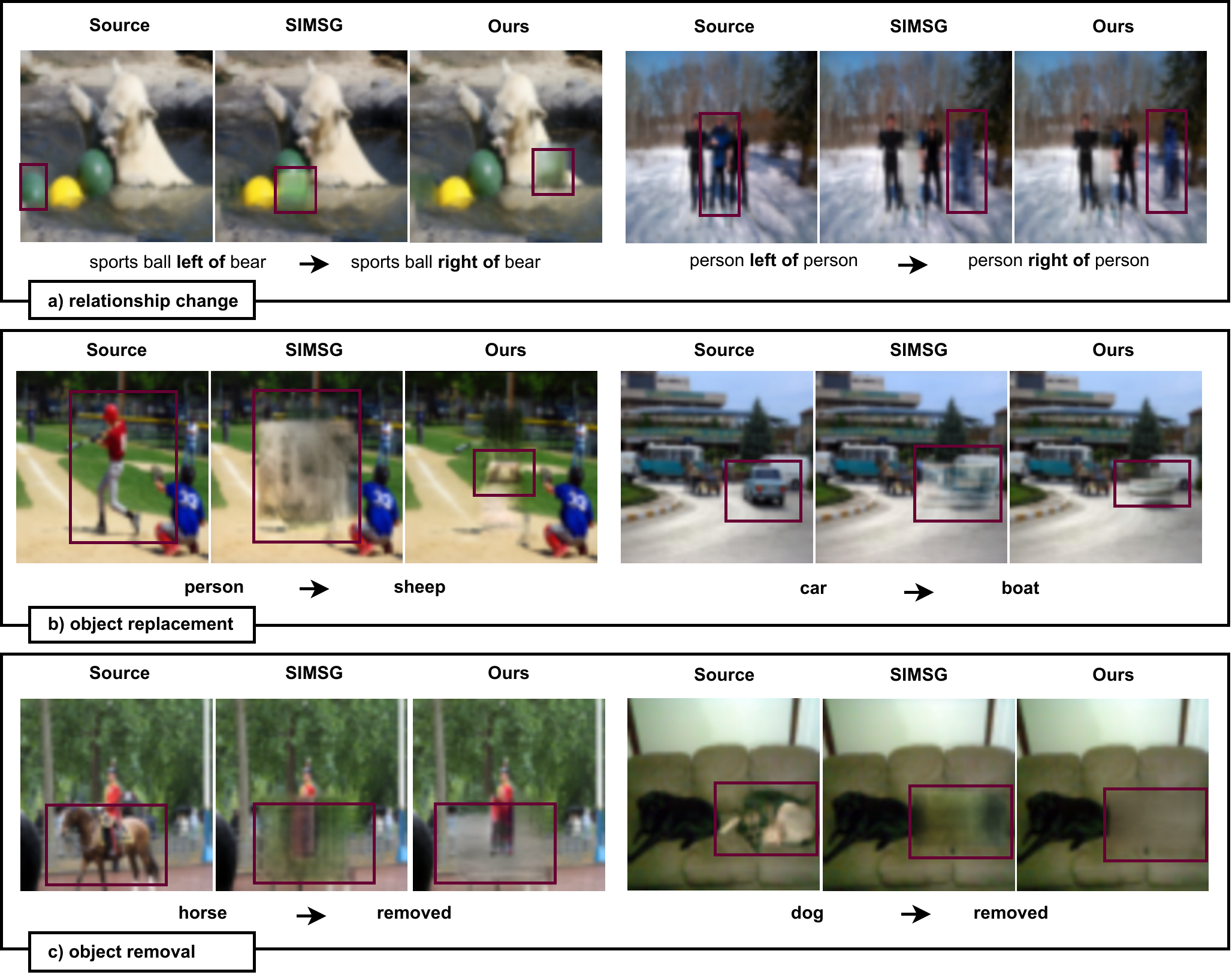}
    \caption{\textbf{Qualitative results for image manipulation on COCO.} Our method shows more accurate positioning and better visual appearance in three main image manipulation tasks of (a) relationship change, (b) object replacement and (c) object removal.}
    \label{fig:qual_res_coco}
\end{figure}

\paragraph{Ablation Study}
The results of our ablation study are reported in \cref{tab:ablation}. First, we present the model performance without the disentanglement. Then, we evaluate the effect of disentangling the latent embeddings. Finally, we show the model performance with disentanglement in both latent embedding and the graph features. Notably, the disentanglement of both components leads to an improvement in most metrics.

\begin{minipage}[c]{0.49\textwidth}
\centering
\small
\captionof{table}{\textbf{Ablation Study on VG}}
\resizebox{0.95\textwidth}{!}{
\begin{tabular}{c@{\hskip 7pt} c@{\hskip 7pt} c @{\hskip 7pt} c@{\hskip 7pt} c @{\hskip 7pt} c@{\hskip 7pt} c}
\toprule
\multicolumn{2}{c}{Disentanglement} & \multicolumn{3}{c}{All pixels} & \multicolumn{2}{c}{RoI only} \\ % \multirow{2}{*}{Method}
 \cmidrule(l{1pt}r{9pt}){3-5} \cmidrule(l{1pt}r{3pt}){6-7}
Embeddings & Graph & \footnotesize{MAE $\downarrow$} & \footnotesize{SSIM $\uparrow$} & \footnotesize{LPIPS $\downarrow$} & \footnotesize{MAE $\downarrow$} & \footnotesize{SSIM $\uparrow$}\\ %
\midrule \midrule
\multicolumn{7}{c}{Generative} \\
\midrule
$-$ & $-$ & $41.88$ & $34.89$ & $0.27$ & N/A & N/A \\
\checkmark & $-$ & $41.80$ & $35.18$ & $0.26$ & N/A & N/A \\
\checkmark & \checkmark & $\bm{41.62}$ & $\bm{35.30}$ & $\bm{0.26}$  & N/A & N/A \\
\midrule \midrule
\multicolumn{7}{c}{GT Graphs} \\
\midrule
$-$ & $-$ & $8.61$ & $87.55$ & $0.050$ & $\bm{21.62}$ & $\bm{58.51}$ \\
\checkmark & $-$  & $8.47$ & $87.53$ & $0.048$ & $21.77$ & $58.30$ \\
\checkmark & \checkmark & $\bm{8.41}$ & $\bm{87.56}$ & $\bm{0.048}$ & $21.76$ & $58.18$ \\
\midrule \midrule
\multicolumn{7}{c}{Predicted Graphs} \\
\midrule
   $-$ & $-$ & $13.82$ & $83.98$ & $0.077$ & $28.82$ & $49.34$ \\
    \checkmark & $-$ & $9.65$ & $86.68$ & $0.054$ & $25.62$ & $51.19$ \\
    \checkmark & \checkmark & $\bm{9.39}$ & $\bm{86.91}$ & $\bm{0.052}$  & $\bm{25.40}$ & $\bm{51.85}$ \\
\bottomrule
\end{tabular}
}
\label{tab:ablation}
\end{minipage}
\begin{minipage}[c]{0.49\textwidth}
\centering
\small
\captionof{table}{\textbf{Image reconstruction on COCO}}
\resizebox{0.95\textwidth}{!}{
\begin{tabular}{l@{\hskip 7pt} c @{\hskip 7pt} c@{\hskip 7pt} c @{\hskip 7pt} c@{\hskip 7pt} c}
\toprule
 \multirow{2}{*}{Method} & \multicolumn{3}{c}{All pixels} & \multicolumn{2}{c}{RoI only} \\
 \cmidrule(l{1pt}r{9pt}){2-4} \cmidrule(l{1pt}r{3pt}){5-6}
 & \footnotesize{MAE $\downarrow$} & \footnotesize{SSIM $\uparrow$} & \footnotesize{LPIPS $\downarrow$} & \footnotesize{MAE $\downarrow$} & \footnotesize{SSIM $\uparrow$}\\ %
\midrule \midrule
\multicolumn{6}{c}{Generative} \\
\midrule
SIMSG \cite{dhamo2020semantic} & $54.03$ & $24.12$ & $0.490$ & N/A & N/A \\
\methodName{} (Ours) & $\bm{51.07}$ & $\bm{26.53}$ & $\bm{0.418}$  & N/A & N/A \\
\midrule \midrule
\multicolumn{6}{c}{Non Generative} \\
\midrule
SIMSG \cite{dhamo2020semantic}  & $9.36$ & $87.00$ & $0.086$ & $27.68$ & $49.93$ \\
\methodName{} (Ours) & $\bm{9.24}$ & $\bm{88.26}$ & $\bm{0.057}$ & $\bm{27.52}$ & $\textbf{50.35}$ \\
\bottomrule
\end{tabular}
}
\label{tab:coco}
\vspace{20pt}
\captionof{table}{\textbf{User study on VG}}
\resizebox{0.95\textwidth}{!}{
    \begin{tabular}{ccccc}
    Method & Removal & Replacement & Relationship Change & Mean \\ \hline
        SIMSG \cite{dhamo2020semantic} & 14.06 & 27.68 & 26.95 & 23.51 \\ \hline
        \methodName{} (Ours) & \textbf{85.94} & \textbf{72.32} & \textbf{73.04} & \textbf{76.49}\\ 
    \end{tabular}
    
    \label{tab:user_study}
    }
%\end{table}
\end{minipage}

\subsection{Discussion}
We showed that our proposed model outperforms the previous works in all scenarios by generating images with higher quality and more meaningful results. We also showed that the generated images by our method have higher diversity and less artifacts. In our user study, the users were given the option to choose which model performs better image manipulation in terms of image quality and how well the change in the image corresponds to the modification in the graph. The \methodName{} model was chosen as the best model compared to SIMSG \cite{dhamo2020semantic} in $76.49\%$ of the cases. Nevertheless, our method has some limitations similar to the related work, which we discuss here.
\paragraph{Limitations}
The dominant limitations of our approach are manipulating high-resolution images and the reconstruction of faces and complex scenes. We believe that these limitations originate from the difficulty in generating high-quality images from scene graphs \cite{johnson2018image} that could be due to the wild nature of images in the VG dataset and occasional errors in the scene graph annotations. We assume that, it would be possible to overcome this issue by using a higher quality dataset with scene graphs and semantic segmentation annotations. Regarding the face reconstruction problem, although this is an easy task given datasets of pure face images, the model fails to generalize well to the faces when combined with images in the wild. We present some failure case examples in the supplementary material.

\section{Conclusions}
\label{sec:conc}
We presented a novel disentangling framework for image manipulation using scene graphs. The results of our experiments showed that using disentangled representation in the latent embedding for semantic image manipulation is an effective way to improve image generation and manipulation quality. The variational representation for object features enables generating diverse images compared to previous work. Further, we showed that using a disentangled graph neural network for extracting the scene graph features provides more meaningful and useful features for the disentangled latent embedding, resulting in higher reconstruction performance. As a future direction, we consider improving the decoder network by taking advantage of diffusion models.
% ---- Bibliography ----

\bibliography{main}
\end{document}